\documentclass{article}

\usepackage{arxiv}
\usepackage[utf8]{inputenc}
\usepackage[T1]{fontenc}
\usepackage{amsmath,amsfonts}
\usepackage{graphicx}
\usepackage{float}
\usepackage{booktabs}
\usepackage{url}
\usepackage{cite}

\title{Cost Optimization in Production Line Using Genetic Algorithm}

\author{
  Alireza Rezaee \\
  Department of Mechatronics, School of Intelligent Systems\\
  College of Interdisciplinary Science and Technology\\
  University of Tehran\\
  Tehran, Iran\\
  \texttt{arrezaee@ut.ac.ir} \\
}

\begin{document}
\maketitle

\begin{abstract}
This paper presents a genetic algorithm (GA) approach to cost‐optimal task scheduling in a production line. The system consists of a set of serial processing tasks, each with a given duration, unit execution cost, and precedence constraints, which must be assigned to an unlimited number of stations subject to a per-station duration bound. The objective is to minimize the total production cost, modeled as a station-wise function of task costs and the duration bound, while strictly satisfying all prerequisite and capacity constraints. Two chromosome encoding strategies are investigated: a station-based representation implemented using the JGAP library with SuperGene validity checks, and a task-based representation in which genes encode station assignments directly. For each encoding, standard GA operators (crossover, mutation, selection, and replacement) are adapted to preserve feasibility and drive the population toward lower-cost schedules. Experimental results on three classes of precedence structures—tightly coupled, loosely coupled, and uncoupled—demonstrate that the task-based encoding yields smoother convergence and more reliable cost minimization than the station-based encoding, particularly when the number of valid schedules is large. The study highlights the advantages of GA over gradient-based and analytical methods for combinatorial scheduling problems, especially in the presence of complex constraints and non-differentiable cost landscapes.
\end{abstract}

\section{Introduction to GA}

Nature has a robust way of evolving successful organisms. The organisms that are ill suited for an environment die off, whereas the ones that are fit live to reproduce. Offspring are similar to their parents, so each new generation has organisms that are similar to the fit members of the previous generation. If the environment changes slowly, the species can gradually evolve along with it, but a sudden change in the environment is likely to wipe out a species. Occasionally, random mutations occur, and although most of these mean a quick death for the mutated individual, some mutations lead to new successful species. The publication of Darwin's \emph{The Origin of Species} on the Basis of Natural Selection was a major turning point in the history of science.

It turns out that what's good for nature is also good for artificial systems \cite{
AbdoliHajati2014,
Ayatollahi2015,
BarolliAINA2024,
BarolliBWCCA2019,
BarolliWAINA2019,
Barzamini2012,
CremersACCV2014,
Fiorini2019,
Hajati2017Surface,
Hajati2006FaceLocalization,
Hajati2010PoseInvariant,
Hajati2017DynamicTexture,
Mahajan2024_ens,
Pakazad2006FaceDetection,
Sadeghi2024COVID_new,
Shojaiee2014Palmprint,
Sopo2021DeFungi,
Tavakolian2022FastCOVID_new,
Tavakolian2023Readmission_new,
Wang2022SoftwareImpacts_new,
KarimiRezaee2017Helmholtz,
MohamadzadeRezaee2017Antenna,
Ramezani2024Drones,
Rezaee2008GeneticSymbiosis,
Rezaee2010FIR,
Rezaee2017PID,
Rezaee2017Penetrometer,
Rezaee2017MPC,
RezaeeGolpayegani2012,
RezaeePajohesh2016,
Sadeghi2024ECG,
Taghvaee2014Metamaterial,
Tavakolian2022SoftwareImpacts_new,
Gavagsaz2018LoadBalancing,
Rezaee2014FuzzyCloud,
Sarvghad2011ThinkingStyles,
Shahramian2013Leptin,
Shahramian2013Troponin
}. The pseudo-code for \texttt{GENETIC-ALGORITHM} starts with a set of one or more individuals and applies selection and reproduction operators to ``evolve'' an individual that is successful, as measured by a fitness function. There are several choices for what the individuals are. They can be entire agent functions, in which case the fitness function is a performance measure or reward function, and the analogy to natural selection is greatest. They can be component functions of an agent, in which case the fitness function is the critic. Or they can be anything at all that can be framed as an optimization problem.

Since the evolutionary process learns an agent function based on occasional rewards (offspring) as supplied by the selection function, it can be seen as a form of reinforcement learning. \texttt{GENETIC-ALGORITHM} simply searches directly in the space of individuals, with the goal of finding one that maximizes the fitness function. The search is parallel because each individual in the population can be seen as a separate search. It is hill climbing because we are making small genetic changes to the individuals and using the best resulting offspring. The key question is how to allocate the searching resources: clearly, we should spend most of our time on the most promising individuals, but if we ignore the low-scoring ones, we risk getting stuck on a local maximum. It can be shown that, under certain assumptions, the genetic algorithm allocates resources in an optimal way.

Usually there are only two main components of most genetic algorithms that are problem dependent, the problem encoding and the evaluation function.

The first step in the implementation of any genetic algorithm is to generate an initial population. As shown in Figure 1, one generation is broken down into a selection phase and recombination phase. This figure shows strings being assigned into adjacent slots during selection. In fact they can be assigned slots randomly in order to shuffle the intermediate population. Mutation (not shown) can be applied after crossover.

\begin{figure}[H]
\centering
\includegraphics[width=0.6\linewidth]{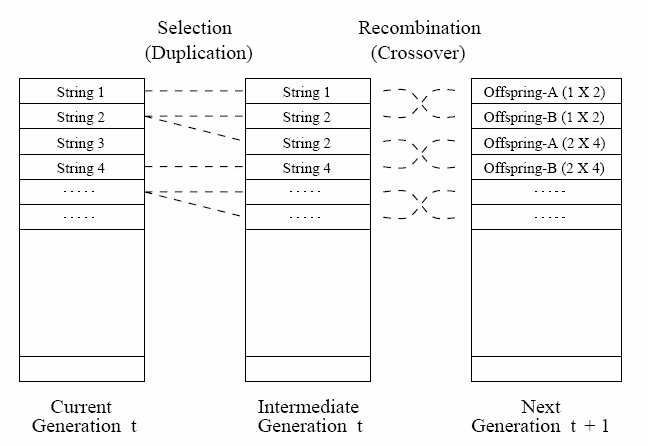}
\caption{The evolution process (schematic).}
\label{fig:evolution}
\end{figure}

\section{Problem Specification}

A production line consists of $N$ processing tasks $T_i$ ($i = 1..N$) that should be done in serial to produce a product. Each task $T_i$ has a duration $D(T_i)$ and a unit cost $C(T_i)$ which indicate task execution duration (in hours) and the execution cost of $T_i$ per hour, respectively. Thus, the total execution cost of $T_i$ can be computed as $D(T_i) \times C(T_i)$. Moreover, some tasks are prerequisite to the others; we show it by $\text{Pre}(T_i, T_j)$ which means that $T_i$ is prerequisite to $T_j$ and should be done before it; it’s obvious that the $\text{Pre}$ relation is transitive which means that $\text{Pre}(A, B)$ and $\text{Pre}(B, C)$ imply $\text{Pre}(A, C)$.

On the other hand, each task should be done in a station along the production line; however, one station may handle more than one task; again in serial. There is infinite number of stations assumed which are placed in serial. Moreover, there is duration bound $K$ for each station which states that the sum of durations of the tasks assigned to each station should not be greater than $K$.

Let
\[
S_j = \{ T_i \mid T_i \text{ is assigned to station } S_j \}
\]
\[
\text{Cost}(S_j) = K \times \max_{T_i \in S_j} C(T_i)
\]
\[
\text{Total Cost} = \sum_j \text{Cost}(S_j)
\]

The problem is to assign the $N$ processing tasks to the stations in a way that the total cost is minimized as well as the prerequisite relations and the duration bound constraint are all satisfied.

\section{Two Approaches to Problem Encoding}

In this section, two solutions for problem encoding are suggested. The first one is known as \emph{station-based} and the second is named \emph{task-based}. The difference between these two solutions is the chromosome structure they suggest for the problem. Implementation of the station-based approach is done with JGAP library while the second approach is completely implemented.

\section{Genome Representation}

\subsection{Station-based}

\subsubsection*{Used JGAP classes}

\paragraph{Gene}

\begin{verbatim}
public interface Gene extends java.lang.Comparable,
                              java.io.Serializable
\end{verbatim}

Genes represent the discrete components of a potential solution (the Chromosome). This interface exists so that custom gene implementations can be easily plugged-in, which can add a great deal of flexibility and convenience for many applications. Note that it's very important that implementations of this interface also implement the equals() method. Without a proper implementation of equals(), some genetic operations will fail to work properly.

\begin{verbatim}
void setAllele(java.lang.Object a_newValue)
\end{verbatim}

Sets the value of this Gene to the new given value. The actual type of the value is implementation-dependent.

\paragraph{IntegerGene}

\begin{verbatim}
public class IntegerGene extends NumberGene implements Gene
\end{verbatim}

A Gene implementation that supports an integer values for its allele. Upper and lower bounds may optionally be provided to restrict the range of legal values allowed by this Gene instance.

\paragraph{SuperGene}

\begin{verbatim}
public interface Supergene extends Gene
\end{verbatim}

Super gene represents several genes, which usually control closely related aspects of the phenotype. The super gene mutates only in such way, that the allele combination remains valid. Mutations, that make allele combination invalid, are rejected inside \texttt{Gene.applyMutation(int, double)} method. Supergene components can also be super genes, creating the tree-like structures in this way.

In biology, the invalid combinations represent completely broken metabolic chains, unbalanced signaling pathways (activator without suppressor) and so on.

At least about 5\% of the randomly generated super gene superallele values should be valid. If the valid combinations represent too small part of all possible combinations, it can take too long to find the suitable mutation that does not break a super gene. If you face this problem, try to split the super gene into several sub-super genes.

In order to have a suitable chromosome for this problem, gene is taken as only have one task. All of chromosomes are composed of the constant set of genes and they are only different in the order that their genes are arranged.

Due to existence of prerequisite relationship among genes of every chromosome, it’s better to use super gene, a pool of genes that automatically checks the accuracy of sought relationship in new chromosomes.

A class named \texttt{SuperGene} is defined that extends \texttt{abstractSupergene} – a JGAP library class – and the \texttt{IsValid()} function of JGAP class is implemented, in order to check chromosomes validity after crossover and/or mutation.

In order to use JGAP standard chromosomes one must pass a valid sample chromosome to \texttt{setSampleChromosome()} function. A chromosome is an array of genes; so in order to use super genes, chromosomes are defined as a temp array of genes with length one and filled up the only index of this array with the defined \texttt{SuperGene} instance. With this trick there would be a chromosome composed of a one length array of genes that its only block is a supergene, filled by integergenes.

\begin{verbatim}
public class SuperGene extends abstractSupergene { … };

SuperGene superGene = new SuperGene(geneArray);
Gene[] temp1 = new Gene[1];
temp1[0] = superGene;
\end{verbatim}

\paragraph{Chromosome}

\begin{verbatim}
public class Chromosome extends java.lang.Object
    implements java.lang.Comparable,
               java.lang.Cloneable,
               java.io.Serializable
\end{verbatim}

Chromosomes represent potential solutions and consist of a fixed-length collection of genes. Each gene represents a discrete part of the solution. Each gene in the Chromosome may be backed by a different concrete implementation of the Gene interface, but all genes in a respective position (locus) must share the same concrete implementation across Chromosomes within a single population (genotype). In other words, gene 1 in a chromosome must share the same concrete implementation as gene 1 in all other chromosomes in the population.

The implementation of chromosome is as follows:

\begin{verbatim}
Gene[] sampleGenes = new Gene[Main.numberOfTasks];

for (int i = 0; i < Main.numberOfTasks; i++)
    sampleGenes[i] = new IntegerGene(0, Main.numberOfTasks - 1);

SuperGene superGene = new SuperGene(sampleGenes);
Gene[] temp1 = new Gene[1];
temp1[0] = superGene;
config.setSampleChromosome(new Chromosome(temp1));
\end{verbatim}

The crossover and mutation functions of JGAP are not overwritten. These operations are entirely devolved to JGAP defined functions. The only process that is handled is checking the validity of the resulted chromosomes after crossover or mutation. This is done by overwriting the \texttt{IsValid()} function of \texttt{abstractSupergene} class of JGAP.

The overwritten \texttt{IsValid()} function is composed of three checking parts:

\begin{itemize}
\item Checking for existence of every gene in chromosome; if not, chromosome is invalid.
\item Checking for duplication of any task in chromosome, if yes, chromosome is invalid.
\item Checking for if a prerequisite task of current task is come after it; if yes chromosome is invalid.
\end{itemize}

The invalid chromosome will be thrown out of chromosome pool.

\subsection{Task-based}

In this solution, gene is simply an integer value. The chromosome has length equal to the number of tasks. The value of element \texttt{gene[i]} in each chromosome is interpreted as the station number which processes task number $i$. This means that indexing is done via task numbers.

There exists a function for checking the validity of the chromosome according to the constraints of the problem. The first constraint is that no station is allowed to contain a set of tasks such that their sum of duration is greater than $K$. This is checked simply by scanning the chromosome once and summing up stations duration, i.e.\ if \texttt{gene[i] = j}, then the duration of task number $i$, is added to the total duration of station $j$. If at any time, station duration overflows $K$, that chromosome is not valid.

The second constraint is the prerequisite satisfaction. The prerequisite condition $\text{Pre}(T_i, T_j)$ is satisfiable whenever the station number assigned $T_i$ is less than or equal to the station number assigned to $T_j$, i.e.\ \texttt{genes[i] <= genes[j]}. Since each task has a list of its prerequisites which is given as input, by scanning this list once, the condition explained above can be checked. So the total cost of checking the validity of the chromosome is $O(n^2)$.

\section{Cross Over Operator}

\subsection{Station-based}

\begin{verbatim}
public class CrossoverOperator extends java.lang.Object
    implements GeneticOperator
\end{verbatim}

The crossover operator randomly selects two Chromosomes from the population and ``mates'' them by randomly picking a gene and then swapping that gene and all subsequent genes between the two Chromosomes. The two modified Chromosomes are then added to the list of candidate Chromosomes. This \texttt{CrossoverOperator} supports both fixed and dynamic crossover rates. A fixed rate is one specified at construction time by the user. This operation is performed $1/m\_crossoverRate$ as many times as there are Chromosomes in the population. A dynamic rate is one determined by this class if no fixed rate is provided.

\subsection{Task-based}

The cross over is implemented in a separate class. The goal is to produce two valid children from the parents given. The function starts with selecting a random cross over point. Then the first slot of the first parent is concatenated with the second slot of the second parent to produce the first child. The symmetric procedure is done for the second child. Since the cost of producing a valid chromosome is too much and all the error checking would be too error-prone to implement, the procedure simply creates a child and then checks for its validity. Invalid child is discarded and the procedure continues until two valid children are produced. The order of this algorithm is therefore $O(n \times x)$, where $n$ is the number of tasks and $x$ is a non-deterministic value showing the number of time the cross over point should be changed, so that a valid child is produced.

\section{Mutation Operator}

\subsection{Station-based}

\begin{verbatim}
public class MutationOperator extends java.lang.Object
    implements GeneticOperator
\end{verbatim}

The mutation operator runs through the genes in each of the Chromosomes in the population and mutates them in statistical accordance to the given mutation rate. Mutated Chromosomes are then added to the list of candidate Chromosomes destined for the natural selection process.

This \texttt{MutationOperator} supports both fixed and dynamic mutation rates. A fixed rate is one specified at construction time by the user. A dynamic rate is determined by this class if no fixed rate is provided, and is calculated based on the size of the Chromosomes in the population such that, on average, one gene will be mutated for every ten Chromosomes processed by this operator.

\subsection{Task-based}

Each child produced in the previous section is mutated with a constant probability. In this method two random indexes in the gene array of the chromosome, are selected. Because mutation does nothing but an unusual change in the chromosome, this function also changes the values of these two selected elements. In the context of the problem this procedure means selecting two tasks and swapping the stations that are assigned to them. Because the swap operation may results in an invalid chromosome, the validity is checked at the end of the procedure and the function wouldn’t return until a valid mutation has taken place. It is worth mentioning that not all the children are mutated. The order of mutation is $O(x)$ where $x$ is the non-deterministic value describing the number of times required to produce a valid mutation.

\section{Fitness Function}

\subsection{Station-based}

The fitness value is computed by assigning as much task as possible to a station. Then the maximum cost of tasks assigned to each station is computed and multiplied by constant $K$. The reverse of this result would be the fitness value of the chromosome.

\subsection{Task-based}

Through scanning the chromosome once, the maximum task cost assigned to each station is computed. These maximum costs are then summed up and multiplied by value $K$. This will produce the total cost of all stations. The fitness value will be $1/\text{total cost}$, so that the generations will move toward producing the least cost.

\section{Selection and Replacement}

\subsection{Station-based}

The JGAP library supports a class for selecting the chromosomes for cross over and mutation in each evolution. There exists a class named \texttt{WeightedRoulette} for a selection algorithm which simply selects chromosome number $i$, with probability $\text{fitness}[i]/\text{total fitness}$. This selector is added to the configuration at the beginning of the configuration settings, so that this method is applied to the generation at the beginning of each evolution.

The program is configured so that the generation size would be constant during several evolutions. There is also another setting for preserving the fittest individual through generations. By these settings, at the end of each evolution the fittest individuals are selected for the next generation and for the rest of generation, weighted roulette is used to choose the next population.

\subsection{Task-based}

Selecting individuals for the cross over operation is done according to their fitness. At first the expression $\text{fitness}[i] / \text{total fitness}$, is computed for each gene in the chromosome. Let’s name this fraction $\text{proportion}[i]$. Next these fraction are added in the way that
\[
\text{proportion}[i] = \left( \sum_{j=1}^{i-1} \text{proportion}[j] \right) + \text{proportion}[i]
\]
It is obvious that the expression above would produce floating point numbers which lead to 1, i.e.\ $\text{proportion}[n] = 1$. Next the algorithm uses a uniform random number generator to produce a floating point number between zero and one. If the number falls in the range of $(\text{proportion}[i], \text{proportion}[i+1])$, then value $i+1$ is selected for the cross over.

It is worth mentioning that the number of times cross over operator is invoked depends on a constant field named \texttt{numberOfCandidateParents}. After the cross over and mutation operation, the population size will grow up to $2 \times \text{numberOfCandidateParents} + \text{initialPopulationSize}$. So in order to keep population size constant and confine individual production, some of the individuals must be deleted. In this phase the algorithm uses a random number generator to produce a random integer value, which is the index of the chromosome to be removed from the population. This chromosome may be either an old parent or a newly generated child.

\section{Results}

The results are generally provided for the task-based solution. Since the fitness evaluation of the first approach doesn’t always lead to the best generation. Although the fitness algorithm in the station-based solution tries to put as much task as possible in a station, this may not produce the appropriate result. Since minimizing the number of stations used is not an issue in this problem. So the first approach doesn’t introduce the desired generations. But as in the task-based approach there is explicit station assignment to each task, the error described above wouldn’t happen.

The results are analyzed in three classes as shown below.

\subsection{Case I: Tightly coupled}

In this case only a single sequence of tasks is allowed to be executed serially. Since the constraints are too restrictive, GA will converge in early generation (before 100) as shown in the average fitness diagram. The prerequisite relation is shown in Figure 2.

\begin{figure}[H]
\centering
\includegraphics[width=0.7\linewidth]{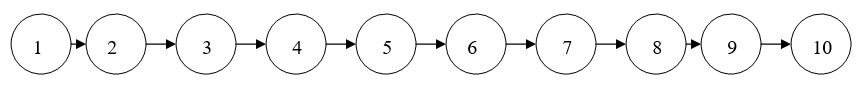}
\caption{Case I, prerequisite diagram.}
\end{figure}

\begin{figure}[H]
\centering
\includegraphics[width=0.7\linewidth]{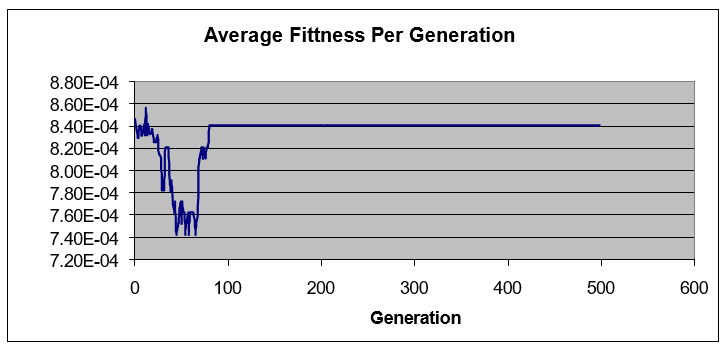}
\caption{Case I, average fitness per generation.}
\end{figure}

\begin{figure}[H]
\centering
\includegraphics[width=0.7\linewidth]{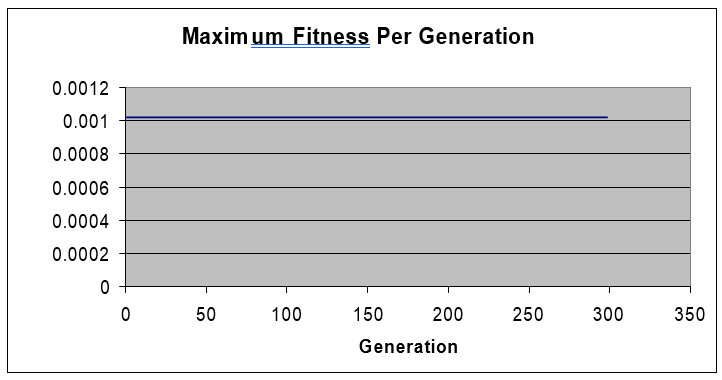}
\caption{Case I, maximum fitness per generation.}
\end{figure}

\begin{figure}[H]
\centering
\includegraphics[width=0.7\linewidth]{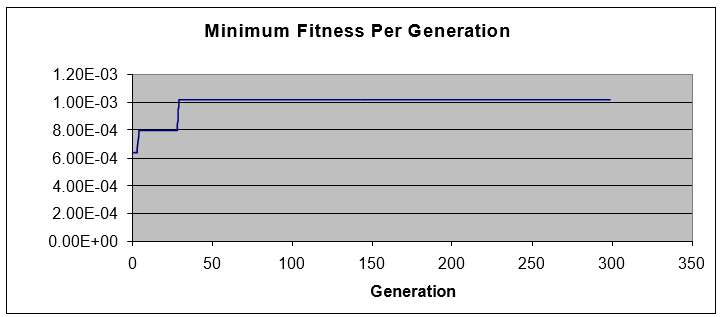}
\caption{Case I, minimum fitness per generation.}
\end{figure}

\subsection{Case II: Loosely coupled}

In this class, there exists some prerequisite relations between tasks, but these relation are some how relaxed, so that the number of possible valid tasks sequences is considerable. As shown in Figure 7, the average fitness has a little fluctuations but using a moving window shows that it is an increasing function as it must logically be. The minimum and maximum fitness diagrams are not as flat as case I.

\begin{figure}[H]
\centering
\includegraphics[width=0.7\linewidth]{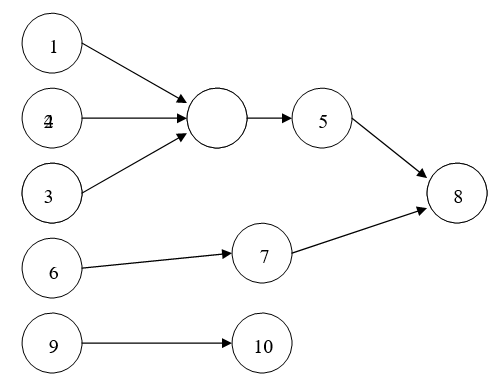}
\caption{Case II, prerequisite diagram.}
\end{figure}

\begin{figure}[H]
\centering
\includegraphics[width=0.7\linewidth]{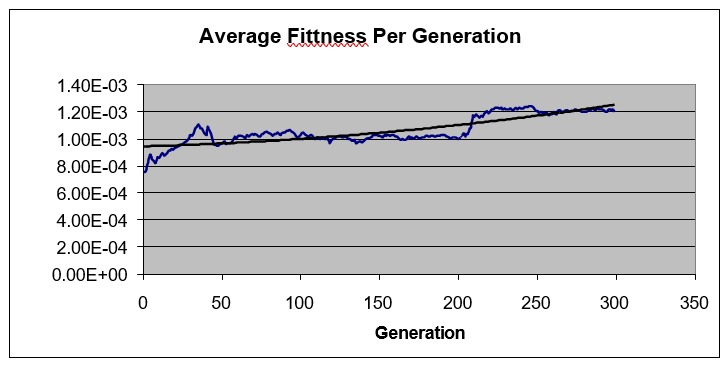}
\caption{Case II, average fitness per generation.}
\end{figure}

\begin{figure}[H]
\centering
\includegraphics[width=0.7\linewidth]{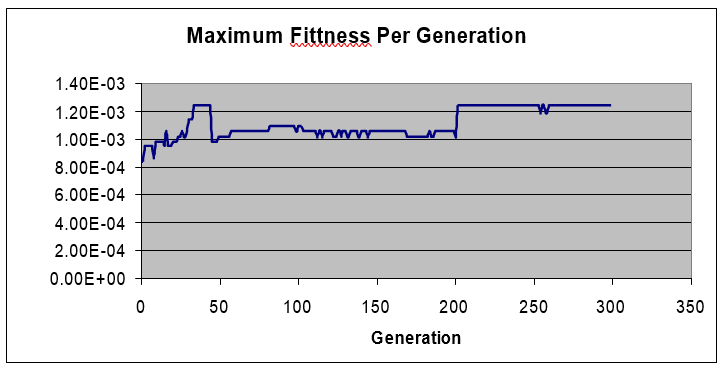}
\caption{Case II, maximum fitness per generation.}
\end{figure}

\begin{figure}[H]
\centering
\includegraphics[width=0.7\linewidth]{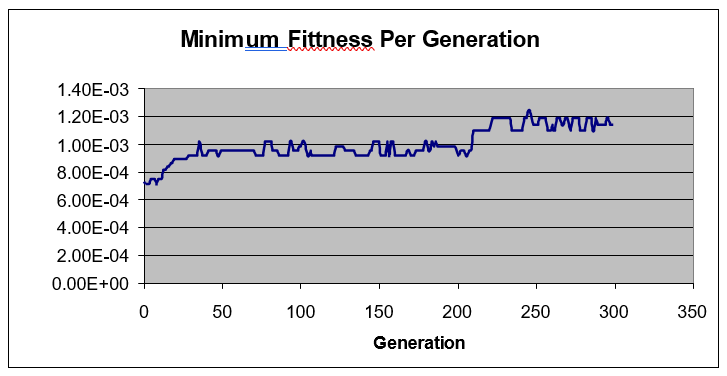}
\caption{Case II, minimum fitness per generation.}
\end{figure}

\subsection{Case III: No coupling}

This class contains test cases with no prerequisite relations. So the only checking during the program will be the maximum total station duration. For this specific problem input, the result from the first approach, station-based, is also brought. But as it is obtainable from Figures 10 and 11, the diagram is smoother in task-based approach which leads to earlier convergence.

\begin{figure}[H]
\centering
\includegraphics[width=0.7\linewidth]{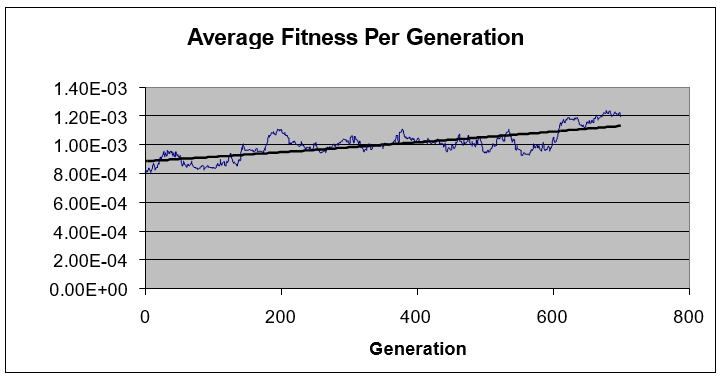}
\caption{Case III, average fitness, task-based.}
\end{figure}

\begin{figure}[H]
\centering
\includegraphics[width=0.7\linewidth]{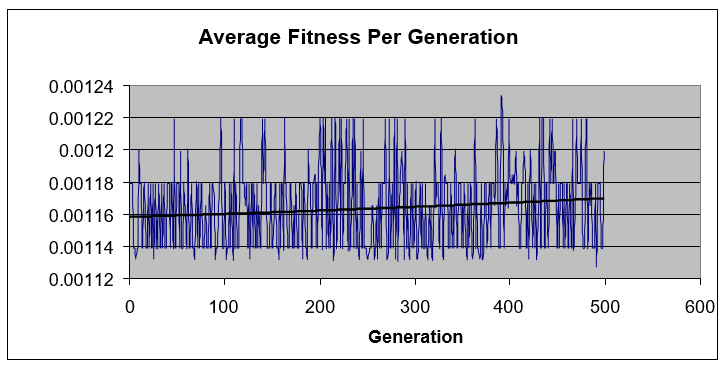}
\caption{Case III, average fitness, station-based.}
\end{figure}

\begin{figure}[H]
\centering
\includegraphics[width=0.7\linewidth]{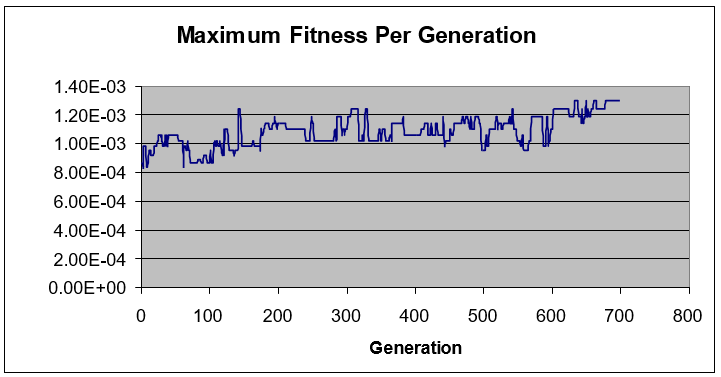}
\caption{Case III, maximum fitness.}
\end{figure}

\begin{figure}[H]
\centering
\includegraphics[width=0.7\linewidth]{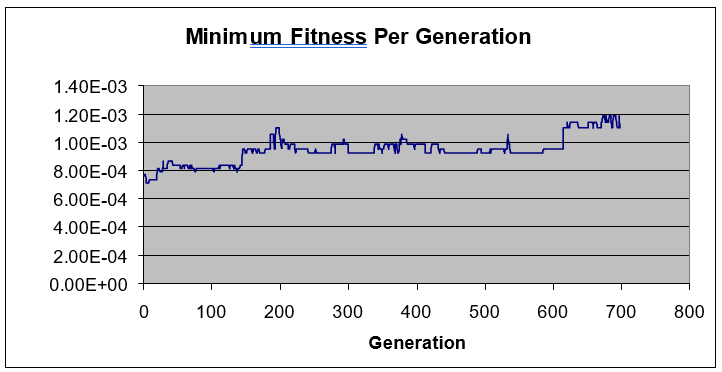}
\caption{Case III, minimum fitness.}
\end{figure}

\section{Optimization techniques}

\subsection{Analytical}

Given $y = f(x)$, take the derivative of $f$ with respect to $x$, set the result to zero, solve for $x$. It works perfectly, but only for simple, analytical functions.

\subsection{Gradient-based (Hill climbing)}

Given $y = f(x)$, pick a point $x_0$, compute the gradient of $f(x_0)$, step along the gradient to obtain $x_1 = x_0 + \alpha f(x_0)$ and repeat this process until extremum is obtained. This approach requires existence of derivatives, and easily gets stuck on local extrema.

\subsection{Enumerative}

Test every point in the state space in order.

\subsection{Random}

Test every point in the state space in order.

\subsection{Genetic Algorithm}

This approach does not require derivatives, but just an evaluation function (a fitness function). It samples the space widely, like an enumerative or random algorithm, but more efficiently. This solution can search multiple peaks in parallel, so is less hampered by local extrema than gradient-based methods. Crossover allows the combination of useful building blocks, or schemata (mutation avoids evolutionary dead-ends) and finally it is robust!

\section{Conclusion}

Through the discussion above, gradient-based techniques may get stuck in local maximum and fail to traverse the whole state space. This error doesn’t occur in GA. Also there exist some heuristics that avoid gradient-based failure in local extrema.
Also in problems where derivations are hard to obtain, hill climbing is not preferable. In this specific problem, applying gradient-based solution has the risk of getting stuck in local maximum and returning a non-optimal task scheduling at last. Whenever good heuristics are used for local maximum avoidance, hill climbing is also applicable to the problem by providing $x$ axis with all the set of chromosomes (all the permutations of genes) and $y$ axis with individual fitness.

\end{document}